\documentclass[journal]{IEEEtran}
\usepackage{booktabs}
\usepackage[scaled=1.0]{helvet}
\usepackage{times}
\usepackage{graphicx}
\usepackage{subfigure}
\usepackage{parskip}
\usepackage{multirow}
\usepackage{amsmath}
\usepackage{color}
\newcommand{\bZ}{\mathbf{Z}}
\usepackage{booktabs}
\usepackage{multirow}
\usepackage{color}
\usepackage{amssymb}
\usepackage{balance}
\usepackage{bm}
\usepackage{cite}
\usepackage{subfigure}
\usepackage[table, xcdraw]{xcolor}
\usepackage{makecell}

\newcommand{\be}{\mathbf{e}}

\usepackage{threeparttable}

\usepackage[labelfont=bf,textfont=it]{caption}
\usepackage[lined,boxed,commentsnumbered, ruled]{algorithm2e}

\begin{document}

\title{Physics Guided Diffusion Model Based  Exposure Interpolation for High Dynamic Range Imaging}
\author{Zhengguo  Li,~\IEEEmembership{Fellow,~IEEE}, and Chaobing Zheng,~\IEEEmembership{Member~IEEE} 
		
		\thanks{ 		Zhengguo Li is with Institute of Advanced Intelligence and Computing, A*STAR, Singapore, 138632, (email: Li\_Zhengguo@a-star.edu.sg).
        Chaobing Zheng is with the school of Information Science and Engineering, Wuhan University of Science and Technology, Wuhan 430081, China (email: zhengchaobing@wust.edu.cn).  
		}
	}

\maketitle

\begin{abstract}
Ghosting artifacts are an open problem for popular multi-shot high dynamic range (HDR) imaging. An alternative solution is to capture two large-exposure-ratio (LER) standard dynamic range (SDR) images simultaneously via a single-shot. Brightness order reversal (BOR) artifacts could be  generated when the two LER images are directly fused together  using a conventional filter based multi-scale exposure fusion (MEF) algorithm. Fortunately, the BOR artifacts can be removed through interpolating an image with an exposure time between those of two LER images.
In this paper, a novel physics-guided diffusion model based algorithm is proposed to interpolate such an image. An image is initially interpolated by a physics-driven exposure interpolation component that is on top of intensity mapping functions, and  residual high-frequency details are generated by a diffusion model based exposure interpolation component. The proposed diffusion model is based on the correlation analysis of generative models and includes a new predicted target to reduce the variation of the Pearson correlation between the noisy data and the target, as well as the amplification factor of the fitting error. Compared with existing neural augmented exposure interpolation algorithms on top of convolution neural networks, the proposed exposure interpolation algorithm has an inherent advantage to generate more high frequency information. Experiment results demonstrate that the proposed exposure interpolation algorithm outperforms existing exposure interpolation algorithms, and the BOR artifacts are removed when the interpolated image and two LER images are fused by a conventional filter based MEF algorithm. 
\end{abstract}


\section{Introduction}
Many real-world scenes such as stadium entrances and nighttime roads
 are with high dynamic ranges (HDRs) while a standard dynamic range (SDR) image can only capture limited dynamic range of these real-world scenes \cite{Debevec97,Mertens07_6}. Details in highlight and shadow regions of these scenes cannot be captured simultaneously by an existing camera via a single-shot. There are over-exposed/under-exposed regions in an SDR image captured by the existing camera, and the saturation could affect subsequent tasks such as object segmentation \cite{xi2023online}, etc. An effective solution is to capture multiple SDR images with different exposures and combine them together to produce an HDR or sRGB image \cite{Debevec97,Mertens07_6}. The differently exposed  SDR images are usually captured sequentially by multiple shots. Both camera movements and moving objects are issues for such HDR imaging systems. It is not difficult to address the former while ghosting artifacts caused by the moving objects is an open problem for multi-shot HDR imaging \cite{1tao2026survey} even though there are many ghost removal algorithms \cite{1hasinoff2016,kalantari2017deep,liu2023unsupervised}.

Two large-exposure-ratio (LER) SDR images can be captured simultaneously by a single-shot using the special devices in \cite{1tanaka2018,1tocci2011}. The beam splitting based system in \cite{1tocci2011} can capture two LER images by a single-shot. Canon released a novel
global shutter with a specific sensor that reads the sensor
twice in an HDR mode \cite{1tanaka2018}. It can also capture two LER images by a single-shot. Ghosting artifacts are not an issue for the single-shot HDR imaging. However, shadow regions in the bright one could be brighter than highlight regions in the dark image. There are brightness order reversal (BOR) artifacts when the two LER images are directly fused together via conventional filter-based multi-scale exposure fusion (MEF) algorithms \cite{Mertens07_6,1kou2017,1wangq2020,1jia2022} even though they are good at preserving the scene depth. Many data-driven single-scale exposure fusion (SEF) algorithms are proposed to address the BOR artifacts  for the fusion of two LER images \cite{1prabhakar2017,xu202DN0fusiondn,wu2024hybrid,huang2024leveraging}. Unfortunately, the data-driven SEF algorithms have a difficulty in preserving the scene depth  of real-world HDR scene in the fused image. An alternative solution is to interpolate an image with an exposure time in between the two LER images \cite{1yang2018} and fuse the two LER images and interpolation image using a conventional filter based MEF algorithm or a data-driven MEF algorithm. It is thus highly demanded to develop an efficient exposure interpolation algorithm for the single-shot HDR imaging on top of two LER images. 

In this paper, a novel explicit exposure interpolation algorithm is proposed through seamless integration of physics-driven and diffusion model-based methods. Similar to  \cite{1yang2018,1zheng2023,1lizg2025}, an image is first explicitly interpolated from the two LER images using a physics-driven exposure interpolation component. Due to the limited representation capability of  the physics-driven exposure interpolation component, there exists residual information between the ground-truth image and the initially interpolated image \cite{1zheng2023,1lizg2025}. The residual image is then explicitly generated via a diffusion model-based exposure interpolation component which is different from the convolution neural network (CNN)-based  components in \cite{1zheng2023,1lizg2025}. Our diffusion model-based exposure interpolation component has an inherent advantage to generate more high-frequency information than the CNN-based exposure interpolation components in \cite{1zheng2023,1lizg2025}. Instead of directly using exiting diffusion models \cite{1ho2020,1rombach2022,1Salimans2021,1gu2022,1Karras2022,1song2023,1liu2022,1lu2024}, a new diffusion model is  proposed on top of theoretical analysis results in \cite{li2026correlation}. The proposed diffusion model has a forward  process and a reverse  process. The forward  process begins with the ground truth of the residual image and evolves to a noise sample by progressively adding the Gaussian noise to the residual image. A predicted target is usually defined in the forward process, and a neural network is trained to map the noisy data to the predicted target during the  forward  process.  The reverse  process starts from the noise sample and converts it back to the corresponding residual image through progressively reducing the Gaussian noise from the noisy data. The noisy data and the output of the neural network at the current moment are utilized to estimate the ground-truth of the residual image and the Gaussian noise in the reverse process. They are adopted to predict the noisy data at the next moment. The reverse process is usually conducted via many steps. Fortunately, the sampling process can be optimized using trajectory distillation  \cite{1Salimans2021}. On the other hand, the fitting error of the neural network could be amplified during the trajectory distillation \cite{1Salimans2021}.  In addition to the well known amplification factor on the fitting error of the neural network, theoretical analysis in \cite{li2026correlation} indicates that the variation of the Pearson correlation between the noisy data and the predicted target is also important to reduce the overall fitting error. Therefore, the predicted target plays an important role in the diffusion model. A new predicted target is proposed in this paper to reduce the variation of the Pearson correlation between the predicted target and the noise data as well as  the amplification factor of the fitting error. The new predicted target outperforms the predicted target in \cite{yue2023resshift}. A new diffusion model is then built up using the new predicted target. Finally, the new model is adopted to design a novel physics-guided diffusion model based  exposure interpolation algorithm for the single-shot HDR imaging on top of two LER images. Experiment results show that the proposed exposure interpolation algorithm outperforms the exposure interpolation algorithms in \cite{1yang2018,1zheng2023,1lizg2025}. The BOR artifacts are indeed removed when the interpolated image and two LER images are fused by the conventional filter based MEF algorithm in \cite{1kou2017}. Overall, the main contributions of this paper are:

1) a new diffusion model is proposed to minimize the variation of the Pearson correlation between the predicted target and the noise data as well as the amplification factor of the fitting error. The new model  outperforms the model in \cite{yue2023resshift};

2) the new diffusion model is  adopted to develop a new neural augmented explicit exposure interpolation algorithm. More high-frequency information is generated by the proposed exposure interpolation algorithms.

The rest of this paper is organized as follows: preliminary knowledge is provided in section \ref{literaturesurvey}.  Details of the proposed algorithm are presented in section \ref{proposedalgorithm}. Experiment results are given in section \ref{experimentresults} to validate the proposed algorithm. Finally, conclusion remarks are provided in section \ref{conculsionremarks}.

\section{Preliminary Knowledge}
\label{literaturesurvey}
In this section, preliminary knowledge on exposure interpolation and diffusion models is presented. The knowledge is useful for the proposed neural augmented exposure interpolation algorithm.

\subsection{Exposure interpolation}

Let $\{Z_i|1\leq i\leq 7\}$ be a set of differently exposed SDR images from an HDR scene. It covers the whole dynamic range of the HDR scene. All these seven SDR images are aligned, and the relationships among their exposure times are 
\begin{align}
\Delta t_1&<\Delta t_2<\cdots<\Delta t_7,\\
\frac{\Delta t_1}{\Delta t_2}&=
\frac{\Delta t_2}{\Delta t_3}=\cdots=\frac{\Delta t_6}{\Delta t_7}.
\end{align}

Clearly, $\bZ_1$ and $\bZ_7$ are a pair of aligned LER images. They can be captured simultaneously for an HDR scene  by a single-shot \cite{1tanaka2018,1tocci2011}. Given the pair of images $Z_1$ and $Z_7$, an image $\hat{Z}_4$ with exposure time as $\Delta t_4$ is interpolated. There exist two different types of algorithm to interpolate the image $\hat{Z}_4$. One type is  physics-driven exposure interpolation \cite{1yang2018}, and the image $\hat{Z}_4$ is interpolated as 
\begin{align}
\hat{Z}_4=f_m(Z_1,Z_7).
\end{align}
As shown in \cite{11li2015}, two images with different exposures can be mapped to each other using intensity mapping functions (IMFs) between them.   The IMFs can be estimated from camera response functions (CRFs). The algorithm in \cite{1yang2018} is on top of IMFs. As pointed out in \cite{1zheng2023,1lizg2025}, there exists residual information between the image $\hat{Z}_4$ and the interpolated image $f_m(Z_1,Z_7)$.

The other type is  neural augmented exposure interpolation, and the image $\hat{Z}_4$ is interpolated as 
\begin{align}
\label{neuralaugmentation}
\hat{Z}_{4,t}=f_m(Z_1,Z_7)+f_{d,t}(Z_1,Z_7).
\end{align}
An initially interpolated image $f_m(Z_1,Z_7)$ is generated using a physics-driven exposure interpolation component, and the residual image is learned using a data-driven component \cite{1zheng2023,1lizg2025}. As shown in \cite{1lizg2025}, the convergence rate of the neural-augmented exposure interpolation algorithm is much faster than that of the data-driven exposure interpolation algorithm if the physics-driven component $f_m(Z_1,Z_7)$ is accurate. Let $Z_4$ be the ground truth image of $\hat{Z}_{4,t}$. The data-driven component in the neural augmented exposure interpolation algorithm is supposed to learn the residual image $(Z_4-f_m(Z_1,Z_7))$ while the data-driven exposure interpolation algorithm is supposed to learn the image $Z_4$.   

The data-driven components in \cite{1zheng2023,1lizg2025} are designed using CNNs. Unfortunately, CNNs have a fundamentally difficulty in learning high-frequency information. A new neural-augmented exposure interpolation algorithm is proposed in this paper using a new diffusion model to replace the CNNs in \cite{1zheng2023,1lizg2025}.

\begin{figure*}[htb!]
\centering
\includegraphics[width=1.0\textwidth]{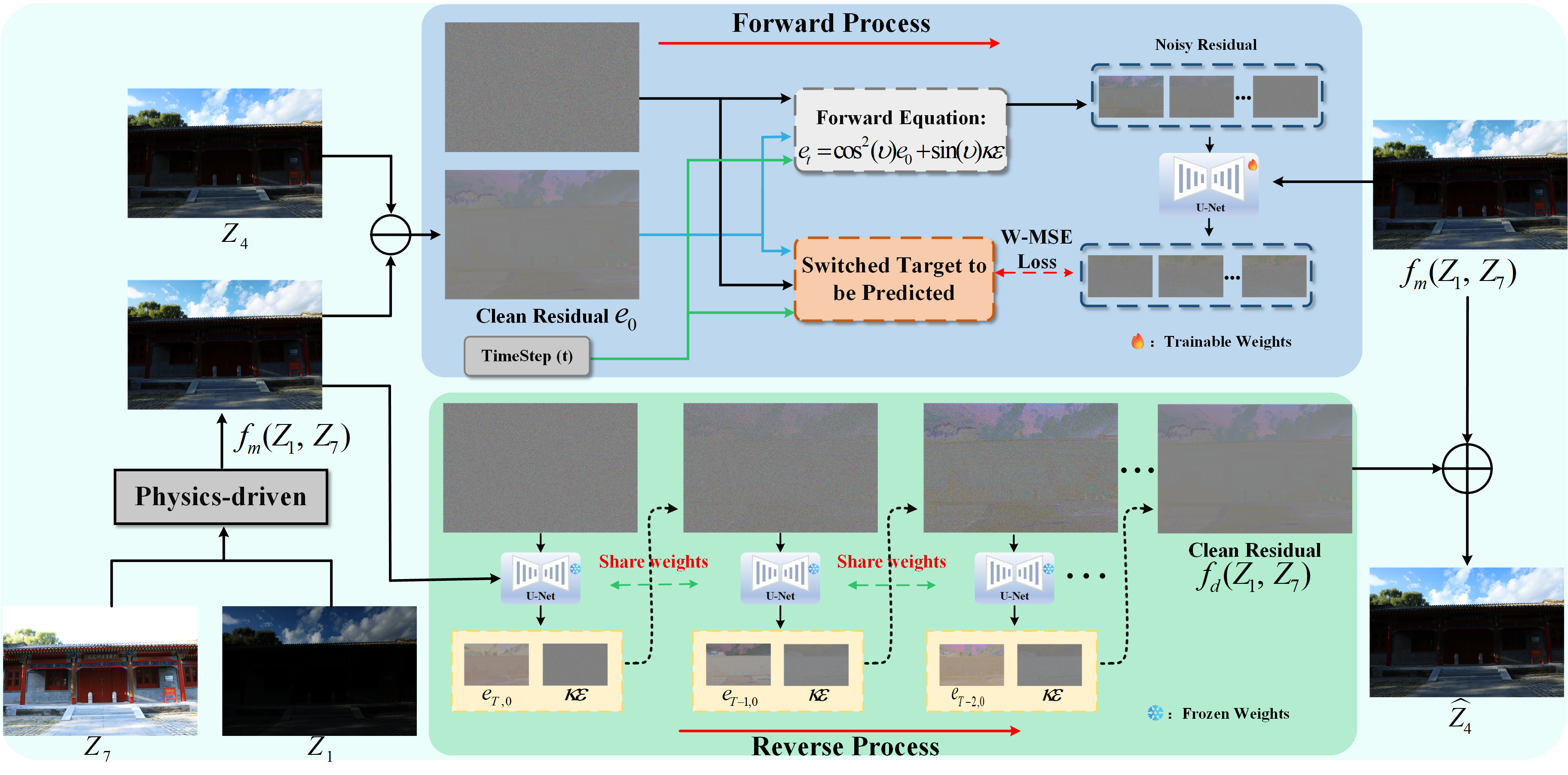}
\caption{Overview of the proposed neural augmented exposure interpolation algorithm on top of a new diffusion model. Through reducing the variation of $\Psi_{\be_t,\omega_t}$ as well as $|\Phi_1(t',t)|$ and $|\Phi_2(t',t)|$, the fitting error of the neural network is reduced in the reverse process. Subsequently, the quality of the interpolated image is improved. }
\label{ProposedFramework}
\end{figure*}

\subsection{Diffusion models}
\label{diffusionmodels}
The goal of generative model is to use a set of data which is sampled from an unknown distribution $p_{d}(\cdot)$ to learn a model for generating new samples from the distribution $p_{d}(\cdot)$. Diffusion models  have emerged as promising and popular frameworks for generative models \cite{1ho2020,1rombach2022,1Salimans2021,1gu2022,1Karras2022,1song2023,1liu2022,1lu2024}. 
Each diffusion model has a forward process and a reverse process. The forward process begins with a ground truth data sample $Z$ and evolves to a noise sample $X_t$ by progressively adding the Gaussian noise $\epsilon$ to the data sample $Z$. The forward process is represented by \cite{li2026correlation}
\begin{align}
X_t=a_{11}(t)Z+a_{12}(t)\epsilon,
\end{align}
where $a_{11}(t)$ and $a_{12}(t)$ specify the forward path from data to noise. Different choices of $a_{11}(t)$ and $a_{12}(t)$ are presented in \cite{1ho2020,1rombach2022,1Salimans2021,1gu2022,1Karras2022,1song2023,1liu2022,1lu2024}.  The choices affect both training efficiency and sampling efficiency.  Let $T$ be the number of steps. The forward process is from 1 to T. The forward time sequence $\Gamma_f$ is defined as \cite{yue2023resshift}
\begin{align}
\label{eq:forwardsequence}
\Gamma_f=\{1,2,\cdots, T\},
\end{align}
$a_{11}(t)$ and $a_{12}(t)$ satisfy 
\begin{align}
\left\{\begin{array}{l}
\lim_{t\rightarrow 1}a_{11}(t)=1\; ;\;
\lim_{t\rightarrow T}a_{11}(t)=0\\
\lim_{t\rightarrow 1}a_{12}(t)=0\; ;\;
\lim_{t\rightarrow T}a_{12}(t)=1
\end{array}
\right..
\end{align}

A neural network $f_{\theta}(X_t,t)$ can be trained to map the noisy data $X_t$ and $t$ to a predicted target $\omega_t$ during the  forward process. The predicted target is defined as \cite{li2026correlation}
\begin{align}
\omega_t=a_{21}(t)Z+a_{22}(t)\epsilon,
\end{align}
where $a_{21}(t)$ and $a_{22}(t)$ are two hyper-parameters. Different choices of $a_{21}(t)$ and $a_{22}(t)$ are also provided in \cite{1ho2020,1rombach2022,1Salimans2021,1gu2022,1Karras2022,1song2023,1liu2022,1lu2024}.
The output of the neural network $f_{\theta}(X_t,t)$ and the noisy data $X_t$ will be utilized to estimate the ground-truth data $Z$ and the Gaussian noise $\epsilon$ in the reverse process. 

The reverse process starts from the noise sample $\epsilon$ and converts it back to the corresponding data sample $Z$ through progressively reducing the Gaussian noise $\epsilon$ from the noisy data $X_t$. The reverse process is from $T$ to 1. Considering two instances $t$ and $t'$ satisfying $t'<t$, the noisy data $X_t$ and the prediction $f_{\theta}(X_t,t)$
are available at $t'$ in the reverse  process. Defining a matrix $A(t)$ as \cite{li2026correlation}
\begin{align}
A(t)=\left[\begin{array}{l l}
		a_{11}(t) & a_{12}(t)\\
		a_{21}(t) & a_{22}(t)\\
	\end{array}
	\right],
\end{align}
$Z$ and $\epsilon$ can be estimated as 
\begin{align}
\label{predictionz}
\left[\begin{array}{l}
Z\\
\epsilon\\
\end{array}
\right]=&
\frac{1}{|A(t)|}\left[\begin{array}{l l}
		a_{22}(t) & -a_{12}(t)\\
		-a_{21}(t) & a_{11}(t)\\
	\end{array}
	\right]\left[\begin{array}{l}
		X_t\\
		f_{\theta}\\
\end{array}
\right],
\end{align}
where $|A(t)|$ is the determinant of matrix $A(t)$. 

Subsequently, the reverse process is  given as
\begin{align}
\nonumber
X_{t'}=&a_{11}(t')Z+a_{12}(t')\epsilon\\\label{reverseprocess}
=&\Phi_1(t',t)X_t+\Phi_2(t',t)f_{\theta}(X_t,t),
\end{align}
where $\Phi_1(t',t)$ and $\Phi_2(t',t)$ are
\begin{align}\left\{\begin{array}{l}
\Phi_1(t',t)=\frac{a_{11}(t')a_{22}(t)-a_{12}(t')a_{21}(t)}{|A(t)|}\\
\Phi_2(t',t)=\frac{a_{11}(t)a_{12}(t')-a_{11}(t')a_{12}(t)}{|A(t)|}
\end{array}
\right.,
\end{align}
and $\Phi_2(t',t)$ is the well known amplification factor on the fitting error of $f_{\theta}(X_t,t)$.

\begin{table*}[htbp]
\caption{Objective comparison between different choices of $\alpha$ in equations (\ref{alpha1})  with $T$ as 5 on \cite{1zheng2023,1xu2022} ($\uparrow$larger is better, $\downarrow$smaller is better). }  
\begin{center}
\label{alpha}
\rowcolors[]{2}{white!20}{gray!15}
	\resizebox{1.75\columnwidth}{!}
	{
		\begin{tabular}{c|c|c|c|c|c|c|c|c}
			\toprule[1.pt]
		$\alpha$ & PSNR$\uparrow$  &  SSIM$\uparrow$    & FSIM$\uparrow$  & LPIPS$\downarrow$   & Entropy$\uparrow$     & CLIPIQA$\uparrow$     & MUSIQ$\uparrow$     & HyperIQA$\uparrow$        \\\hline  
             0.02  &{\bf 33.55}      &0.9489     &{\bf 0.9900}      & {\bf 0.0441}    & {\bf 6.8237}     & 0.4702    & {\bf 63.40}    & {\bf 67.23}      \\
             0.1   &33.44      &{\bf 0.9493}     &0.9899      & 0.0442    & 6.8139     & 0.4681    & 63.28    & 67.13       \\
             1.0   &33.12      &0.9447     &0.9893      & 0.0454    & 6.8212     & {\bf 0.4775}    & 62.77    & 66.91        \\
		\bottomrule[1.pt]
		\end{tabular}
	}
    \end{center}
\end{table*}

The reverse time sequence is defined as 
\begin{align}
\label{eq:ourreversesequence}
\Gamma_r=\{t_1,t_2,\cdots, t_{N_r}\},
\end{align}
where  $t_1$ is $T$ and $t_{N_r}$ is 1. The set $\Gamma_f$ and $\Gamma_r$ can be the same. By repeating the reverse process (\ref{reverseprocess}) from $t_1$ to $t_{N_r}$ along the reverse time sequence $\Gamma_r$ (\ref{eq:ourreversesequence}), $X_{t_{N_r}}$ is obtained from $X_{t_1}(=\epsilon)$ as 
\begin{align}
\nonumber
X_{t_{N_r}}=& \prod_{i=1}^{N_r-1}\Phi_1(t_{i+1},t_i)X_{t_1}+\\\label{eq:newreverse1}
&\sum_{i=1}^{N_r-1}\prod_{j=i+1}^{N_r-1}\Phi_1(t_{j+1},t_j)\Phi_2(t_{i+1},t_i)f_{\theta}(X_{t_i},t_i).
\end{align}

In addition to $|\Phi_1(t',t)|$ and $|\Phi_2(t',t)|$, the overall prediction error also depends on the fitting errors of $f_{\theta}(X_{t_i},t_i)(i=1,\cdots,N_r-1)$ as shown in equation (\ref{eq:newreverse1}). As indicated in \cite{li2026correlation}, the fitting error of $f_{\theta}(X_t,t)$ depends on the following Pearson correlation between the $X_t$ and the $\omega_t$: 
\begin{align}
\Psi_{X_t,\omega_t}=\frac{\mbox{cov}(X_t,\omega_t)}{\sigma_{X_t}\sigma_{\omega_t}},
\end{align}
where $\mbox{cov}(X_t,\omega_t)$ is the covariance of the noisy data $X_t$ and the predicted target $\omega_t$. $\sigma_{X_t}$ and $\sigma_{\omega_t}$ are the standard deviation of $X_t$ and $\omega_t$, respectively.

Since the neural network $f_\theta(\cdot)$ is shared among all $t$'s, the fitting error of  $f_\theta(\cdot)$ depends on those $t$'s with small $\Psi_{X_t,\omega_t}$'s. Therefore,  it is important to reduce the variation of $\Psi_{X_t,\omega_t}$. Certainly, it is also important to reduce $|\Phi_1(t',t)|$ and $|\Phi_2(t',t)|$. They are the guidelines for the proposed diffusion model in this paper.

The diffusion model has a potential to generate high-frequency information which is missed by the physics-driven and CNN-based methods. A new diffusion model will be presented in this paper to generate high-frequency information that is missing from  the physics-driven exposure interpolation component.

\section{The Proposed  Exposure Interpolation Algorithm}
\label{proposedalgorithm}

$\bZ_1$ and $\bZ_7$ are the inputs of the proposed exposure interpolation algorithm.  Similarly to the neural augmentation framework in \cite{1zheng2023,1lizg2025}, the image $\hat{\bZ}_{4,t}$ is interpolated as in equation (\ref{neuralaugmentation}) with $\hat{\bZ}_{4,0}$ as $\bZ_4$. The physics-driven representation $f_m(\bZ_1,\bZ_7)$ is kept the same as \cite{1lizg2025}. For simplicity, a residual image $\be_t$ is defined as 
\begin{align}
\label{eq:NewresidualImage}
\be_t=\hat{\bZ}_{4,t}-f_m(\bZ_1,\bZ_7).
\end{align}
It follows that 
\begin{align}
\label{eq:residual}
\be_0 = \bZ_4 - f_m(\bZ_1,\bZ_7).
\end{align}
In other words, $\be_0$ is the residual image between the physics-driven component $f_m(\bZ_1,\bZ_7)$ and  the ground-truth (or captured) image $\bZ_4$. Generally, the residual image $\be_0$ contains mainly the high-frequency details that are missing in the  the physics-driven component $f_m(\bZ_1,\bZ_7)$. The conditional distribution $q(\be_0 |  f_m(\bZ_1,\bZ_7))$ is  learned through an efficient diffusion model because the diffusion model has more chance to learn the missing high-frequency information than the CNN-based methods in \cite{1zheng2023,1lizg2025}. 

To save space, $\cos\nu$ and $\sin\nu$ are respectively denoted as $C_{\nu}$ and $S_{\nu}$ in the rest of this paper. 
The forward process of the proposed diffusion model  is defined as  
\begin{align}
\label{eq:forward_markov}
\be_t=a_{11}(t)\be_0+a_{12}(t)\epsilon
=C_{\nu}^2\be_0+S_{\nu}\epsilon,
\end{align}
where $\nu$ is $\frac{\pi t}{2 T}$, and $t\in \Gamma_f$. A deep  neural network with parameters $\theta$, $g_{\theta}(\be_t,f_m,t)$ is trained to map the noisy data $\be_t$ to a predicted target $\omega_t$ during the  forward process. The predicted target is an important component of the diffusion model.
Two different choices of $\omega_t$ are presented as below.

\subsection{The Proposed $\omega_t^{(1)}$}
Inspired by the preliminary knowledge in Section \ref{diffusionmodels}, the target to be predicted, $\omega_t^{(1)}$ is proposed as
\begin{align}
\label{eq:newtarget}
\omega_t^{(1)}=a_{22}^{(1)}(t)\be_0+a_{22}^{(1)}(t)\epsilon
=(\gamma-S_{\nu})\be_0+\epsilon,
\end{align}
where  $\gamma(<1)$ is defined as
\begin{align}
\label{alpha1}
\gamma=\min\{\alpha(C_{\nu}-C_{\frac{\pi\tau}{2T}}), 0\},
\end{align}
$\tau(\in [0,T])$ and $\alpha$ are two hyperparameter. $\tau$ is selected as $T/5$ in this paper.  Three values, $\alpha \in \{0.02, 0.1, 1.0\}$, are examined. As shown in Table \ref{alpha}, the choice of $0.02$ achieves the best results on six
of the eight metrics, including PSNR, FSIM, LPIPS, Entropy, MUSIQ and
HyperIQA. Thus, $\alpha$ is selected as 0.02 in this paper. 

A deep neural network with parameters $\theta$, $g_{\theta}^{(1)}(\be_t,f_m,t)$ is trained to map the noisy data $\be_t$ to the predicted target $\omega_t^{(1)}$ during the  forward process as demonstrated in Algorithm \ref{training}. The Swin-UNet in \cite{cao2022swin} is adopted as the backbone network. In particular, $g_{\theta}^{(1)}(\be_t,f_m,t)$ is obtained as 
\begin{align}
\label{eq:ourtraining}
\min_{\theta}\{\sum_{t\in \Gamma_f}\|g_{\theta}^{(1)}(\be_t,f_m,t)-\omega_t^{(1)}\|_2^2\}.
\end{align}

\begin{algorithm}[t]
\SetKwInOut{Input}{Input}
\SetKwInOut{Output}{Output}
\caption{Explicit Exposure Interpolation via Physics-guided Diffusion Model (Training)}
\label{training}
\Input{Dual LER images $Z_1$ and $Z_7$; Ground-truth middle-exposure image $Z_4$; Total timesteps $T$; Hyperparameters $\alpha, \tau$.}
\Output{Optimized parameters $\theta$ of the neural network $g_\theta^{(1)}$.}

\BlankLine
Compute the physics-driven prior image $f_m(Z_1, Z_7)$ using Intensity Mapping Functions (IMFs)\;
Calculate the high-frequency target residual $e_0 = Z_4 - f_m(Z_1, Z_7)$\;

\Repeat{converged}{
    Sample a timestep $t \sim \text{Uniform}(\{1, 2, \dots, T\})$\;
    Sample Gaussian noise $\boldsymbol{\epsilon} \sim \mathcal{N}(\mathbf{0}, \mathbf{I})$\;
    Calculate angles $\nu = \frac{\pi t}{2T}$, $C_\nu = \cos(\nu)$, and $S_\nu = \sin(\nu)$\;
    Calculate forward noisy residual state via $e_t = C_\nu^2 e_0 + S_\nu \boldsymbol{\epsilon}$\;
    Compute network prediction as $g_\theta^{(1)}(e_t, f_m(Z_1, Z_7), t)$\;
    Calculate a dynamic parameter as $\gamma = \min\{\alpha (C_\nu - C_{\frac{\pi \tau}{2T}}), 0\}$\;
    Define the predicted target as $\omega_t^{(1)} = (\gamma - S_\nu) e_0 + \boldsymbol{\epsilon}$\;
    Update parameters $\theta$ via gradient step on the MSE loss:
    $$\mathcal{L}(\theta) = \left\| g_\theta^{(1)}(e_t, f_m(Z_1, Z_7), t) - \omega_t^{(1)} \right\|_2^2$$
}
\end{algorithm}

The output of the neural network $g_{\theta}^{(1)}(\be_t,f_m,t)$ and the noisy data $\be_t$ will be utilized to estimate the residual image $\be_0$ and the noise $\epsilon$ in the reverse process. It can be derived from equations (\ref{eq:forward_markov}) and (\ref{eq:newtarget}) that 
$\be_0$ and  $\epsilon$ can be estimated as 
\begin{align}
\left\{\begin{array}{l}
\be_0=\frac{1}{1-\gamma S_{\nu}}\be_t-\frac{S_{\nu}}{1-\gamma S_{\nu}}g_{\theta}^{(1)}(\be_t,f_m,t)\\
\epsilon=\frac{S_{\nu}-\gamma}{1-\gamma S_{\nu}}\be_t+\frac{C^2_{\nu}}{1-\gamma S_{\nu}}g_{\theta}^{(1)}(\be_t,f_m,t)
\end{array}
\right..
\end{align}

 As shown in Algorithm \ref{sampling}, the reverse process starts from the noise sample $\epsilon$ and converts it back to the corresponding residual image $\be_0$ through progressively reducing the noise from the noisy data $\be_{t}$. Mathematically, the reverse process is  
\begin{align}
\nonumber
\be_{t'}=&C^2_{\nu'}\be_0+S_{\nu'} \epsilon\\\label{eq:reverse}
=& \Phi_1^{(1)}(t',t)\be_t+\Phi_2^{(1)}(t',t)g_{\theta}^{(1)}(\be_t,f_m,t),
\end{align}
where $t'<t$, $\nu'$ is $\frac{\pi t'}{2 T}$,  $\Phi_1^{(1)}(t',t)$ and $\Phi_2^{(1)}(t',t)$ are two coefficients that are calculated as 
\begin{align}
\label{eq:ourphis}
\left\{\begin{array}{l}
\Phi_1^{(1)}(t',t)=\frac{S_{\nu'}S_{\nu}+C_{\nu'}^2-\gamma S_{\nu'}}{1-\gamma S_{\nu}}>0\\
\Phi_2^{(1)}(t',t)=\frac{S_{\nu'}C_{\nu}^2-C_{\nu'}^2S_{\nu}}{1-\gamma S_{\nu}}<0
\end{array}
\right..
\end{align}

The correlation between $\be_t$ and $\omega_t^{(1)}$ is 
 \begin{align}
 \label{eq:newcorralation2}
\Psi_{\be_t,\omega_t^{(1)}}=\frac{\gamma C^2_{\nu}+S^3_{\nu}}{\sqrt{(C_{\nu}^4+S_{\nu}^2)(1+(\gamma-S_{\nu})^2)}}>0.
\end{align}

It can be verified that 
\begin{align}
\left\{\begin{array}{l}
\frac{d\Phi_1^{(1)}(t',t)}{d\gamma}=\frac{(S_{\nu}-S_{\nu'})C_{\nu}^2}{(1-\gamma S_{\nu})^2}>0\\
\frac{d\Phi_2^{(1)}(t',t)}{d\gamma}=\frac{S_{\nu}(S_{\nu'}-S_{\nu})(1+S_{\nu'}S_{\nu})}{(1-\gamma S_{\nu})^2}<0\\
\frac{d\Psi_{\be_t,\omega_t^{(1)}}}{d\gamma}=\frac{1-\gamma S_{\nu}}{\sqrt{(C_{\nu}^4+S_{\nu}^2)(1+(\gamma-S_{\nu})^2)^3}}>0
\end{array}
\right..
\end{align}

\begin{figure*}[htb!]
\centering
\includegraphics[width=1.0\textwidth]{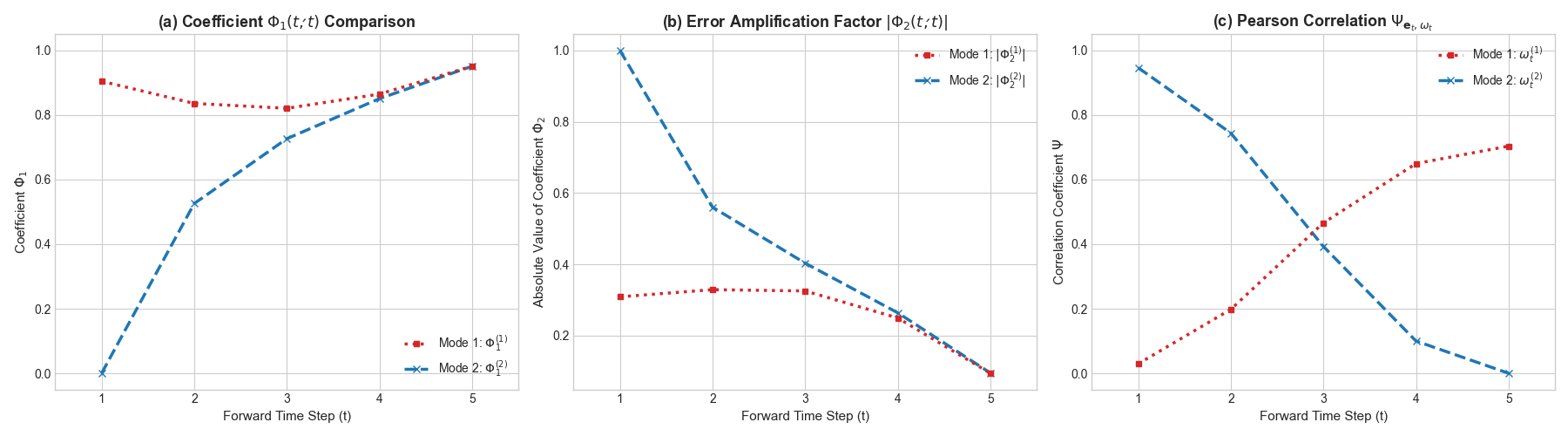}
\caption{$|\Phi_1(t',t)|$, $|\Phi_2(t',t)|$ and $\Psi_{\be_t,\omega_t}$ of two different choices of $\omega_t$ in equations (\ref{eq:newtarget}) and (\ref{eq:NTUTarget}) with $T$ as 5. $|\Phi_2^{(1)}(t',t)|$ is smaller than $|\Phi_2^{(1)}(t',t)|$ and the variation of $\Psi_{\be_t,\omega_t^{(1)}}$ is smaller than the variation of $\Psi_{\be_t,\omega_t^{(2)}}$. }
\label{ComparisonCorrelationFigure}
\end{figure*}

Clearly, $|\Phi_1^{(1)}(t',t)|$, $|\Phi_2^{(1)}(t',t)|$, and $\Psi_{\be_t,\omega_t^{(1)}}$  decrease if $t$ is larger than $\tau$ as shown in Fig. \ref{ComparisonCorrelationFigure}.  Subsequently, the variation of $\Psi_{\be_t,\omega_t^{(1)}}$ reduces. 

\begin{algorithm}[t]
\SetKwInOut{Input}{Input}
\SetKwInOut{Output}{Output}
\caption{Explicit Exposure Interpolation via Physics-guided Diffusion Model (Inference)}
\label{sampling}
\Input{Two LER images $Z_1$ and $Z_7$; Trained network $g_\theta^{(1)}$; Reverse time sequence $\Gamma_r = \{t_1, t_2, \dots, t_{N_r}\}$ with $t_1 = T, t_{N_r} = 1$; Hyperparameters $\alpha, \tau$.}
\Output{Interpolated middle-exposure image $\hat{Z}_4$.}

\BlankLine
Initially interpolates $f_m(Z_1, Z_7)$ using a physics-driven component\;
Initialize noisy state at time $t_1$: $e_{t_1} \sim \mathcal{N}(\mathbf{0}, \mathbf{I})$\;

\For{$i = 1$ \KwTo $N_r - 1$}{
    Set current timestep $t = t_i$ and next timestep $t' = t_{i+1}$ ($t' < t$)\;
    Compute $\nu = \frac{\pi t}{2T}$, $\nu' = \frac{\pi t'}{2T}$, $C_\nu = \cos(\nu)$, $S_\nu = \sin(\nu)$, $C_{\nu'} = \cos(\nu')$, $S_{\nu'} = \sin(\nu')$\;
    Calculate $\gamma = \min\{\alpha (C_\nu - C_{\frac{\pi \tau}{2T}}), 0\}$\;
    Predict target $\hat{\omega}_t^{(1)} = g_\theta^{(1)}(e_t, f_m(Z_1, Z_7), t)$\;
    Compute coefficient $\Phi_1^{(1)}(t', t) = \frac{S_{\nu'} S_\nu + C_{\nu'}^2 - \gamma S_{\nu'}}{1 - \gamma S_\nu}$\;
    Compute coefficient $\Phi_2^{(1)}(t', t) = \frac{S_{\nu'} C_\nu^2 - C_{\nu'}^2 S_\nu}{1 - \gamma S_\nu}$\;
    Update residual state for step $t'$: 
    $$e_{t'} = \Phi_1^{(1)}(t', t) e_t + \Phi_2^{(1)}(t', t) \hat{\omega}_t^{(1)}$$
}
Set final predicted residual $\hat{e}_0 = e_{t_{N_r}}$\;
Reconstruct the target middle-exposure image: $\hat{Z}_4 = f_m(Z_1, Z_7) + \hat{e}_0$\;
\Return $\hat{Z}_4$\;
\end{algorithm}

By repeating the reverse process (\ref{eq:reverse}) from $t_1$ to $t_{N_r}$ along the reverse time sequence $\Gamma_r$ (\ref{eq:ourreversesequence}), $\be_{t_{N_r}}$ is obtained from $\be_{t_1}(=\epsilon)$ as 
\begin{align}
\nonumber
&\be_{t_{N_r}}= \prod_{i=1}^{N_r-1}\Phi_1^{(1)}(t_{i+1},t_i)\be_{t_1}+\\\label{eq:reverse1}
&\sum_{i=1}^{N_r-1}\prod_{j=i+1}^{N_r-1}\Phi_1^{(1)}(t_{j+1},t_j)\Phi_2^{(1)}(t_{i+1},t_i)g_{\theta}^{(1)}(\be_{t_i},f_m,t_i).
\end{align}

Similar to the ResShift \cite{yue2023resshift}, the proposed  diffusion model can be  transferred into  the latent space \cite{esserVQGAN2021,1rombach2022}. The image can be compressed by a factor of four in spatial dimensions to reduce the computational cost.  This can be implemented by  substituting $\bZ_4$ and $f_m(\bZ_1,\bZ_7)$ with their latent codes.

\subsection{Comparison with an Existing $\omega_t^{(2)}$}

In this subsection, the proposed $\omega_t^{(1)}$ in equation (\ref{eq:newtarget}) is compared with the $\omega_t^{(2)}$  in \cite{yue2023resshift} that is defined as 
\begin{align}
\label{eq:NTUTarget}
\omega_t^{(2)}=\be_0\; ;\; t\in \Gamma_f.
\end{align}

The forward process is given in equation (\ref{eq:forward_markov}), and the reverse process is  
\begin{align}
\label{eq:ntureverse}
\be_{t'}= \Phi_1^{(2)}(t',t)\be_t+\Phi_2^{(2)}(t',t)g_{\theta}^{(2)}(\be_t,f_m,t),
\end{align}
where $g_{\theta}^{(2)}(\be_t,f_m,t)$ is obtained as 
\begin{align}
\label{eq:ntutraining}
\min_{\theta}\{\sum_{t\in \Gamma_f}\|g_{\theta}^{(2)}(\be_t,f_m,t)-\omega_t^{(2)}\|_2^2\},
\end{align}
$\Phi_1^{(2)}(t',t)$ and $\Phi_2^{(2)}(t',t)$  are  
\begin{align}
\label{eq:NTUphis}
\left\{\begin{array}{l}
\Phi_1^{(2)}(t',t)=\frac{S_{\nu'}}{S_{\nu}}\\
\Phi_2^{(2)}(t',t)=\frac{C^2_{\nu'}S_{\nu}-S_{\nu'}C^2_{\nu}}{S_{\nu}}
\end{array}
\right..
\end{align}

 The correlation between $\be_t$ and $\omega_t^{(2)}$ is 
\begin{align}
\label{eq:NTUcorrelation}
\Psi_{\be_t,\omega_t^{(2)}}=\frac{C^2_{\nu}}{\sqrt{C^4_{\nu}+S^2_{\nu}}}.
\end{align}

$\omega_t^{(1)} $in equation (\ref{eq:newtarget}) and $\omega_t^{(2)}$ (\ref{eq:NTUTarget}) are now compared. As shown in Fig. \ref{ComparisonCorrelationFigure}, the variation of $\Psi_{\be_t,\omega_t^{(2)}}$ is larger than that of $\Psi_{\be_t,\omega_t^{(1)}}$. $|\Phi_2^{(2)}(t',t)|$ is usually  larger than $|\Phi_2^{(1)}(t',t)|$. 

\section{Experiments}
\label{experimentresults}

The main contribution of this paper is an explicit exposure interpolation algorithm on top of physics-guided diffusion model. Thus,  experiments focus on comparing different exposure interpolation algorithms in this section.

\subsection{Experimental Settings}
Experiments are conducted to validate the effectiveness of the proposed exposure interpolation algorithm. The HDR imaging datasets introduced in \cite{1zheng2023,1xu2022} are adopted for training and evaluation. Specifically, the dataset is divided into three subsets, including 420 sequences for training, 40 sequences for validation, and 60 sequences for testing. To ensure data consistency, camera motion and object movement are strictly controlled, and sequences containing noticeable misalignment are excluded.Data  In addition, all images are randomly cropped into patches of size $512\times512$ to augment the training data and improve model generalization. Furthermore, the dataset from \cite{Cai2018} is utilized solely for testing to evaluate the network's performance, with no involvement in the training stage. During training, the batch size is set to $4$.  The initial learning rate is $5\times10^{-5}$, which is gradually decreased following a cosine annealing schedule. All experiments are implemented in PyTorch and conducted on an NVIDIA A6000 GPU.

\begin{figure*}[htb!]
\centering
\includegraphics[width=1\textwidth]{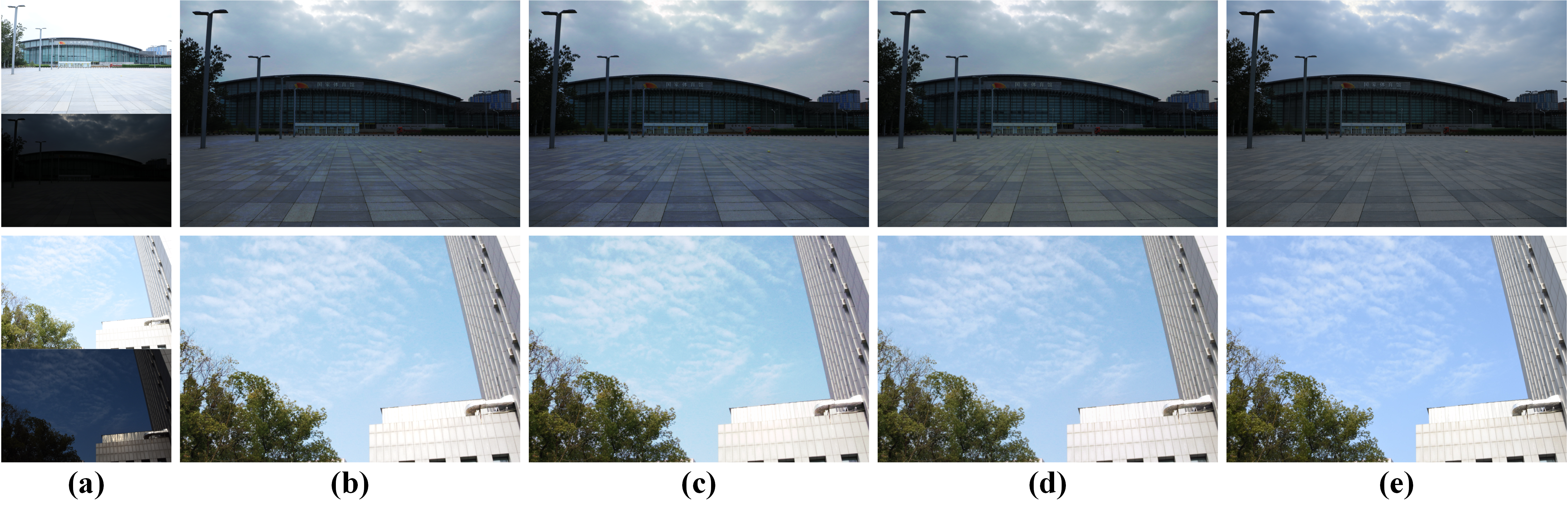}
\caption{Comparison of different physics-driven exposure interpolation algorithms. (a) are the inputs, (b), (c) and (d) show the results obtained by the physics-driven exposure interpolation algorithms in \cite{1yang2018,1zheng2023}, and our adopted physics-driven component, respectively. (e) represents the ground truth.}
\label{myComparisonTarget}
\end{figure*}

\begin{table*}[htbp]
\caption{Objective comparison among the physics-driven exposure interpolation algorithms in \cite{1yang2018, 1zheng2023} and our adopted physics-driven exposure interpolation component ($\uparrow$larger is better, $\downarrow$smaller is better).} 
\label{tabexposureinterpolation}
    \rowcolors[]{2}{white!20}{gray!15}
	\resizebox{2.\columnwidth}{!}
	{
		\begin{tabular}{c|c|c|c|c|c|c|c|c|c}
			\toprule[1pt]
		Methods & Data	& PSNR$\uparrow$  &  SSIM$\uparrow$ & FSIM$\uparrow$   & LPIPS$\downarrow$   & Entropy$\uparrow$     & CLIPIQA$\uparrow$     & MUSIQ$\uparrow$   & HyperIQA$\uparrow$     \\\hline
             Algorithm in \cite{1yang2018}       & \cite{1zheng2023,1xu2022}  & 26.65     &0.9114    &0.9761   &0.0905  &6.4661  &0.4834   &58.996  &65.07   \\
             Algorithm in \cite{1zheng2023} & \cite{1zheng2023,1xu2022}       & 28.07     &0.9142    &0.98   &0.0901  &6.5903  &{\bf 0.4836}   &58.438   &65.38  \\
             Our & \cite{1zheng2023,1xu2022}                                  & {\bf 29.79}     & {\bf 0.919}    &{\bf 0.9823}   &{\bf 0.0848}  &{\bf 6.691}  & 0.4727   &{\bf 59.0875}   &{\bf 65.46} \\\hline
             Algorithm in \cite{1yang2018}       & \cite{Cai2018}       & 24.4     &0.8633    &0.9519   &0.1268  &7.0101  &0.4972   &61.7405  &65.36   \\
             Algorithm in \cite{1zheng2023} & \cite{Cai2018}       & 24.93     &0.8702    &0.957   &0.1192  &7.0443  &0.496   &61.8327  &65.44    \\
             Our & \cite{Cai2018}                                       & {\bf 25.87}     &{\bf 0.8848}    &{\bf 0.9631}   &{\bf 0.1083}  &{\bf 7.0817}  &{\bf 0.4998}   &{\bf 61.8573}    &{\bf 65.53}    \\
		\bottomrule[1.2pt]
		\end{tabular}
	}
\end{table*}

\begin{figure*}[htb!]
\centering
\includegraphics[width=1.0\textwidth]{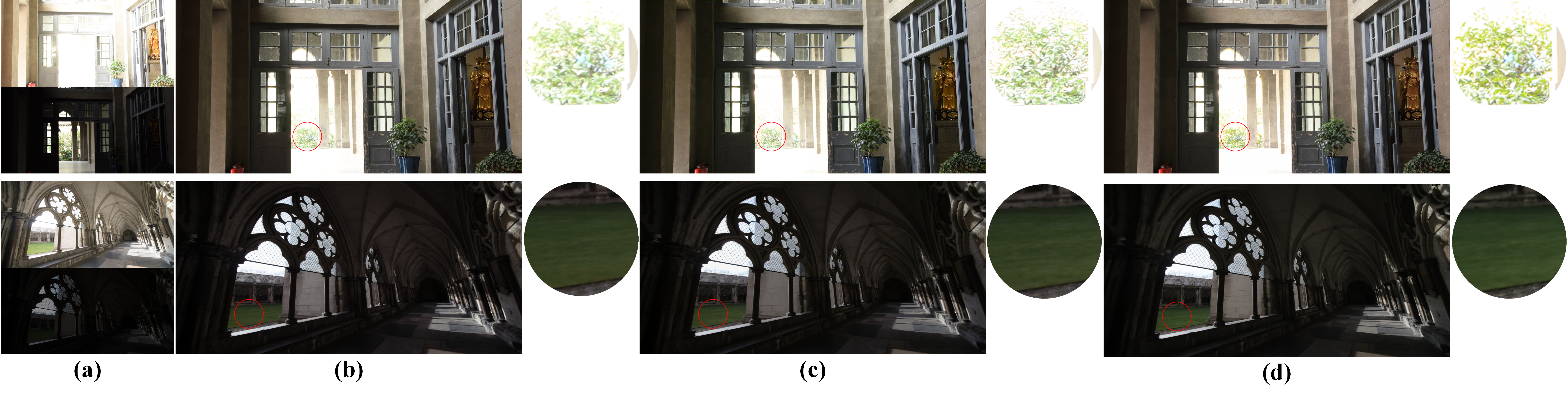}
\caption{Comparison of neural augmented exposure interpolation algorithms in \cite{1zheng2023, 1lizg2025} and the proposed neural augmented exposure interpolation algorithm. (a) Input images with different exposures. (b) Result of \cite{1zheng2023}. (c) Result of \cite{1lizg2025}. (d) Result of the proposed algorithm. The red circles mark the regions shown enlarged on the right of each result. The proposed algorithm generates more high-frequency information that the algorithms in \cite{1zheng2023, 1lizg2025}. }
\label{NeuralResult}
\end{figure*}

\subsection{Comparison of Different Physics-driven Exposure Interpolation Components}
Physics-driven exposure interpolation component plays an important role in the  proposed neural augmented framework. Therefore, in this subsection, we first conduct a comparative evaluation of the adopted physics-driven exposure interpolation component against two representative physics-driven exposure interpolation methods reported in \cite{1yang2018,1zheng2023}. As summarized in Table \ref{tabexposureinterpolation}, the adopted physics-driven component consistently outperforms the competing approaches across multiple widely adopted image quality metrics, including full-reference metrics such as SSIM \cite{wang2004image}, PSNR, FSIM \cite{zhang2011fsim}, and LPIPS \cite{zhang2018unreasonable}, as well as no-reference metrics, including CLIP-IQA \cite{wang2023exploring}, Entropy, MUSIQ \cite{ke2021musiq}, and HyperIQA \cite{su2020blindly}. The interpolated images by our adopted component also approach closer to the ground truth images subjectively as demonstrated in Fig. \ref{myComparisonTarget}. The results of the interpolation methods reported in \cite{1yang2018,1zheng2023} exhibit noticeable color deviation, with a bluish cast in the pavement and building regions of the first scene and washed-out, desaturated skies in the second. In contrast, our result  preserves natural color tones and fine textures such as cloud structures and tile patterns, and is closest to the ground truth

\begin{table*}[htbp]
\caption{Objective comparison of neural augmented exposure interpolation algorithms in \cite{1zheng2023, 1lizg2025} and the proposed neural augmented exposure interpolation algorithm ($\uparrow$larger is better, $\downarrow$smaller is better).} 
	\label{neural}
    \rowcolors[]{2}{white!20}{gray!15}
	\resizebox{2.\columnwidth}{!}
	{
		\begin{tabular}{c|c|c|c|c|c|c|c|c|c}
			\toprule[1pt]
		Methods & Data	& PSNR$\uparrow$  &  SSIM$\uparrow$   & FSIM$\uparrow$    & LPIPS$\downarrow$   & Entropy$\uparrow$     & CLIPIQA$\uparrow$     & MUSIQ$\uparrow$   & HyperIQA$\uparrow$     \\\hline
             Algorithm in \cite{1zheng2023}  & \cite{1zheng2023,1xu2022}  & 32.07     &0.9362    &0.9859   &0.0527  &6.7896  &0.4541   &58.80   &66.19   \\
             Algorithm in \cite{1lizg2025}   & \cite{1zheng2023,1xu2022}  & 33.41     &0.9443    &0.9894   &0.0472  &6.8137  &0.4648   &60.42   &66.75  \\
             Our & \cite{1zheng2023,1xu2022}                              &{\bf 33.55}      &{\bf 0.9489}     &{\bf 0.9900}      & {\bf 0.0441}    & {\bf 6.8237}     & {\bf 0.4702}    & {\bf 63.40}    & {\bf 67.23}  \\ \hline
             Algorithm in \cite{1zheng2023}       & \cite{Cai2018}        & 25.38     &0.8822    &0.9582   &0.1239  &7.1253  &0.4455   &60.64   &65.51   \\
             Algorithm in \cite{1lizg2025} & \cite{Cai2018}               & 26.62     &0.8830    &0.9613   &0.1228  &7.1333  &0.4744   &61.39   &66.36  \\
             Our & \cite{Cai2018}                                         & {\bf 26.71}     &{\bf 0.8922}    &{\bf 0.9621}   &{\bf 0.1207}  &{\bf 7.1406}  &{\bf 0.4745}   &{\bf 64.26}        &{\bf 67.17} \\
		\bottomrule[1.2pt]
		\end{tabular}
	}
\end{table*}

\subsection{Comparison with Existing Neural Augmented Exposure Interpolation Algorithms}
The proposed physics guided diffusion model-based exposure interpolation algorithm is compared with the neural augmented exposure interpolation algorithms in \cite{1zheng2023, 1lizg2025}. As shown in Fig. \ref{NeuralResult}, the proposed diffusion model-based exposure interpolation method produces sharper results with more clearly defined fine details and improved local structural clarity than the CNN-based exposure interpolation algorithms in \cite{1zheng2023, 1lizg2025}. These improvements stem from the progressive generative refinement process, in which the model iteratively reconstructs image content to recover fine-scale structures and enhance local contrast. Compared with the CNN-based exposure interpolation methods, the proposed algorithm better preserves subtle textures and spatial details, yielding visually sharper and more coherent results, particularly in challenging regions with large exposure differences.

These three algorithms are also objectively compared. The results in
Table~\ref{neural} show that the proposed algorithm also slightly outperforms the CNN-based exposure interpolation algorithms in \cite{1zheng2023, 1lizg2025} on both datasets \cite{1zheng2023,1xu2022} and \cite{Cai2018} from all eight metrics points of view .

\begin{table*}[htbp]
\caption{Objective comparison between two different choices of $\omega_t$ in equations (\ref{eq:newtarget})(the proposed one) and (\ref{eq:NTUTarget}) (one in \cite{yue2023resshift}) with $T$ as 5 using date sets in \cite{1zheng2023,1xu2022} ($\uparrow$larger is better, $\downarrow$smaller is better). }  
\label{tabexposureinterpolation}
\begin{center}
\rowcolors[]{2}{white!20}{gray!15}
	\resizebox{1.75\columnwidth}{!}
	{
		\begin{tabular}{c|c|c|c|c|c|c|c|c}
			\toprule[1.pt]
		Target & PSNR$\uparrow$  &  SSIM$\uparrow$    & FSIM$\uparrow$  & LPIPS$\downarrow$   & Entropy$\uparrow$     & CLIPIQA$\uparrow$     & MUSIQ $\uparrow$     & HyperIQA $\uparrow$        \\\hline  
             Eq. (\ref{eq:newtarget})  & {\bf 33.55}     & {\bf 0.9489}    & {\bf 0.9900}    & {\bf 0.0441}   & {\bf 6.8237}   & 0.4702    &  {\bf 63.40}    & {\bf 67.23}         \\
             Eq. (\ref{eq:NTUTarget})  & 32.77     & 0.9429    & 0.9883    &  0.0492  & 6.8181   &  {\bf 0.4704}   &  61.89    & 66.94     \\
		\bottomrule[1.pt]
		\end{tabular}
	}
    \end{center}
\end{table*}

\subsection{Comparison Between Two Physics-guided Diffusion Model-based Exposure Interpolation Algorithms}
\begin{figure}[htb!]
\centering
\includegraphics[width=0.5\textwidth]{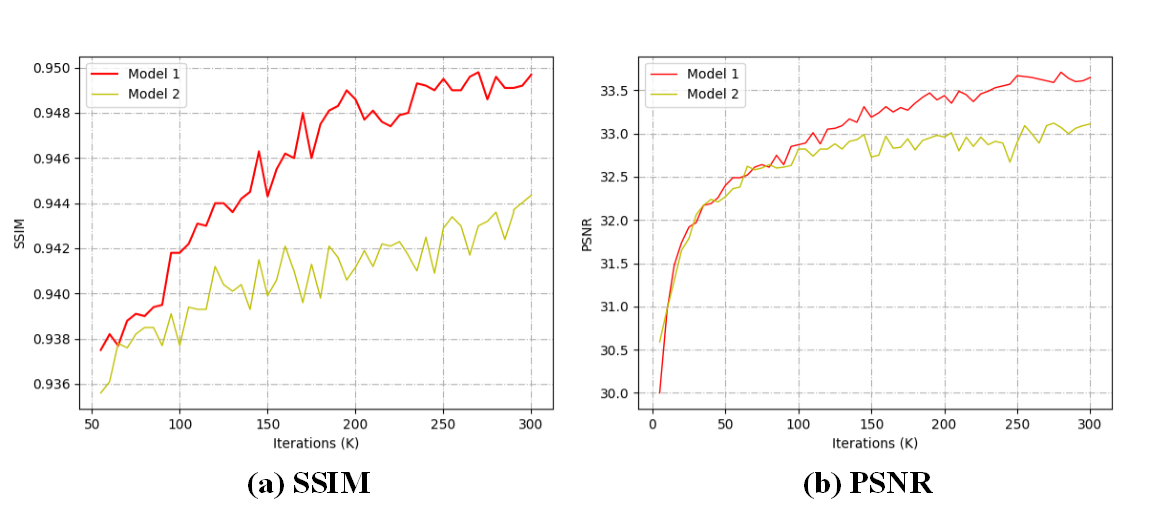}
\caption{Comparison of training curves for two different choices of $\omega_t$ in equations (\ref{eq:newtarget}) and (\ref{eq:NTUTarget}) with $T$ as 5. }
\label{ComparisonTarget}
\end{figure}

The proposed  $\omega_t^{(1)}$ in equation (\ref{eq:newtarget}) is compared with  $\omega_t^{(2)}$ in equation (\ref{eq:NTUTarget}) in this subsection. For simplicity, the simple case with the forward and reverse processes having 5 steps is considered. $\|g_{\theta}(\be_t,f_m,t)-\omega_t\|$  heavily depends on $\|g_{\theta}(\be_{t'},f_m,t')-\omega_{t'}\|$'s with small $\Psi_{\be_{t'},\omega_{t'}}$'s. It is thus important to reduce the variation of  $\Psi_{\be_{t'},\omega_{t'}}$'s. It is also important to reduce $|\Phi_2(t,t')|$. The $\omega_t^{(1)}$ in equation (\ref{eq:newtarget}) is proposed according to this guideline. $|\Phi_2^{(1)}(t,t')|$ is smaller than $|\Phi_2^{(2)}(t,t')|$ as demonstrated in Fig. \ref{ComparisonCorrelationFigure}. Subsequently, the proposed $\omega_t^{(1)}$  in equation (\ref{eq:newtarget}) outperforms the $\omega_t^{(2)}$ in equation (\ref{eq:NTUTarget}) as shown in  Table \ref{tabexposureinterpolation}. Fig. \ref{ComparisonTarget} illustrates the convergence behavior of the two models in terms of SSIM and PSNR at the sane training iterations. Both models improve rapidly during the early stage and gradually stabilize, indicating stable training.  The proposed  $\omega_t^{(1)}$ in equation (\ref{eq:newtarget}) consistently outperforms the $\omega_t^{(2)}$ in equation (\ref{eq:NTUTarget}) as training proceeds. 

\begin{table*}[htbp]
\caption{Objective comparison among the resultant MEF algorithm with 6 state-of-the-art exposure fusion algorithms with inputs as two LER images $Z_1$ and $Z_7$  ($\uparrow$ larger is better).   
\label{tabfusion6EV}
}\rowcolors[]{2}{white!20}{gray!15}
\begin{center}
\resizebox{2.0\columnwidth}{!}
{
\begin{tabular}{c|c|c|c|c|c|c|c|c}
\toprule[1.5pt]
Metrics & Data	 &  GGIF \cite{1kou2017}  &MSF  \cite{1yang2018}  &DeepFuse \cite{1prabhakar2017}   &MEFNet \cite{1ma2019}     &UltraFusion \cite{chen2025ultrafusion} & DiFuser \cite{1tanglf2026}  & Proposed \\\hline
MEF-SSIM$\uparrow$ & \cite{1zheng2023,1xu2022}     &0.8717  & 0.8782     & 0.8579    &0.8662          &0.8834  &0.7806  &{\bf 0.9172}   \\\hline
TMQI$\uparrow$ & \cite{1zheng2023,1xu2022}     &0.8328  & 0.8448     & 0.8219    &0.8185          &0.8365  &0.8507  &{\bf 0.8768}   \\\hline
MEF-SSIM$\uparrow$ & \cite{Cai2018}   &0.9126   & 0.9277  & 0.9104  &0.9010         &0.9409   & 0.8081 &  {\bf 0.9484}   \\\hline
PSNR$\uparrow$ & \cite{Cai2018}       &14.48    & 18.25    & 16.23    &13.30         &20.24     & 15.56 &   {\bf 21.94}   \\\hline
SSIM$\uparrow$ & \cite{Cai2018}   &0.7755   & 0.8119   & 0.7679   &0.7429        &0.8335    & 0.6884 &  {\bf 0.8404}   \\
		\bottomrule[1.2pt]
		\end{tabular}
	}
    \end{center}
\end{table*}

\begin{figure*}[htb!]
\centering
\includegraphics[width=1.0\textwidth]{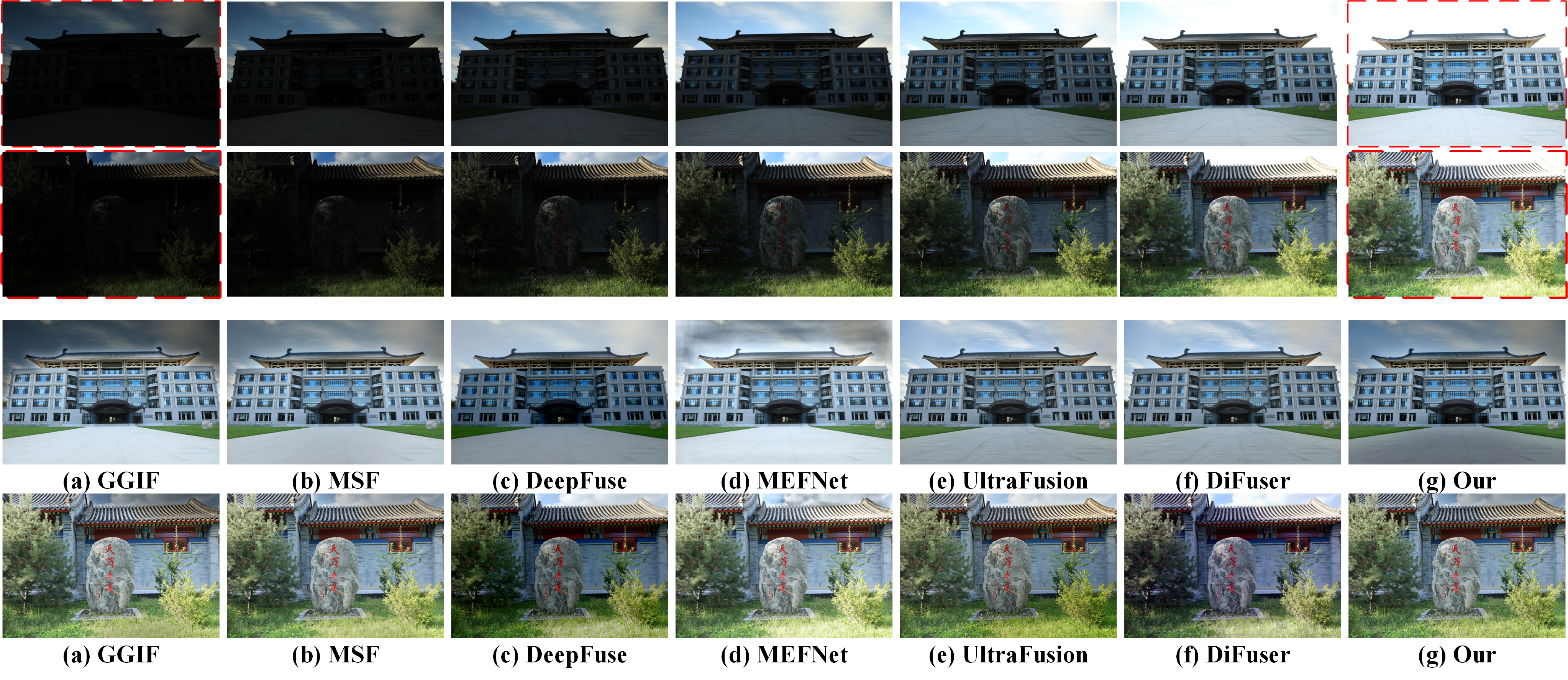}
\caption{Subjective comparison of seven exposure fusion algorithms. These two sets of differently exposed images are from the data set in  \cite{1zheng2023,1xu2022}. All images in the set are given in the top two rows.  The GGIF \cite{1kou2017} suffers from the BOR  artifacts while our resultant MEF algorithm removes the BOR artifacts due to the proposed exposure interpolation algorithm.  }
\label{Fig5Our}
\end{figure*}

\subsection{Application of The Proposed Algorithm to MEF}
To verify the effectiveness of the proposed exposure interpolation, the virtual image $\hat{Z}_4$ and the two LER images $Z_1$ and $Z_7$ are fused together using GGIF \cite{1kou2017}. This is because GGIF in \cite{1kou2017} is the most frequently selected among the thirteen conventional exposure fusion algorithms in \cite{Cai2018}, and it could also outperform  data-driven MEF algorithms if the test sequences and training data have different statistics.  The resultant MEF algorithm is compared with the original   GGIF \cite{1kou2017} and MSF \cite{1yang2018} as well as four state-of-the-art data-driven exposure fusion algorithms including   DeepFuse \cite{1prabhakar2017}, MEFNET \cite{1ma2019}, UltraFusion \cite{chen2025ultrafusion}, and DiFuser \cite{1tanglf2026}.
 The inputs of the GGIF,   DeepFuse, MEFNET, UltraFusion, and DiFuser are the two LER images $Z_1$ and $Z_7$. 
 
 All seven algorithms are first objectively compared.  The MEF-SSIM  \cite{ma20151} for each HDR scene is calculated using the fused image $Z_F$ and the set $\{Z_1, Z_2, Z_3, Z_4, Z_5, Z_6, Z_7\}$. An HDR image is synthesized from the  set $\{Z_1, Z_2, Z_3, Z_4, Z_5, Z_6, Z_7\}$ using the algorithm in \cite{Debevec97}. The TQMI is calculated using the HDR image and the fused image $Z_F$. Since the set $\{Z_1, Z_2, Z_3, Z_4, Z_5, Z_6, Z_7\}$ covers the whole dynamic range of the HDR scene, MEF-SSIM \cite{ma20151} and TQMI \cite{yeganeh2012objective}  measure the similarity between the fused image $Z_F$ and the real world HDR scene. Since ground truth sRGB images are available for the data set in \cite{Cai2018}, the seven algorithms are also compared using PSNR and SSIM. It should be pointed out that the ground truth image is obtained by fusing all the images in the set $\{Z_1, Z_2, Z_3, Z_4, Z_5, Z_6, Z_7\}$ using 13 exposure fusion algorithms including \cite{Mertens07_6,1kou2017}. This implies that PSNR and SSIM also measure the similarity between the fused image $Z_F$ and the HDR scene for the data in \cite{Cai2018}. However, the TMQI is not applicable to the date in \cite{Cai2018} because neither HDR images nor the exposure times of the SDR images are available for all sequences in \cite{Cai2018}. As demonstrated in Table \ref{tabfusion6EV},   the resultant MEF algorithm on top of the proposed exposure interpolation algorithm outperforms the other six exposure fusion algorithms from the MEF-SSIM, TMQI, SSIM and PSNR points of view. The consistent improvements across both datasets indicate that the proposed exposure interpolation can provide useful intermediate-exposure information beyond the two original LER inputs, thereby facilitating the generation of fused images with improved structural fidelity. It should be noted that MSF \cite{1yang2018} outperforms all data-driven exposure fusion algorithms, except UltraFusion \cite{chen2025ultrafusion}. However, MSF \cite{1yang2018} is much faster than UltraFusion \cite{chen2025ultrafusion}.

All seven algorithms are then subjectively compared. Readers are encouraged to zoom in on the electronic version of the figure to more clearly observe the differences among all images. As shown in the top scene of Fig.~\ref{Fig5Our}, both GGIF \cite{1kou2017} and MEFNET \cite{1ma2019} suffer from the BOR artifacts in the sense that the floor is darker than the sky in the real world HDR scene but it is brighter than the sky in the fused images. The BOR artifacts are removed by  MSF \cite{1yang2018} and our resultant MEF algorithm due to the corresponding exposure interpolation algorithms. DeepFuse \cite{1prabhakar2017} has a difficulty in preserving fine details in the fused images.  UltraFusion \cite{chen2025ultrafusion}  suffers from color distortion in the bottom scene.   DiFuser \cite{1tanglf2026} does not preserve fine details in the sky region of the bottom scene. All these problems are also well addressed by our algorithm. However, our algorithm could suffer from halo artifacts because conventional filter based MEF algorithms have a risk to produce halo artifacts.

\section{Conclusion Remarks}
\label{conculsionremarks}
This paper has presented a novel neural augmented explicit exposure interpolation algorithm
for two aligned LER images from a real world HDR scene. A physics-driven component first explicitly 
interpolates an initial image, and a new diffusion model based component then explicitly 
generates the residual high-frequency details. In the proposed diffusion model,  a new predicted target is designed to reduce both the variation of the Pearson correlation between the noisy data and the target and the amplification factor of the fitting error.
 The proposed exposure interpolation algorithm outperforms existing exposure interpolation algorithms, with particularly clear gains in
perceptual metrics such as MUSIQ, and recovers sharper textures in highlight
regions. When the proposed exposure interpolation algorithm is integrated with a conventional filter based MEF algorithm, the resultant MEF algorithm indeed avoids  the BOR artifacts from appearing in the fused image. It also outperforms six state-of-the-art exposure fusion algorithms.  However, the speed of our exposure interpolation algorithm is an issue. Future work will focus
on reducing the sampling cost of the reverse process.

\bibliographystyle{IEEEtran}
\bibliography{main}

\end{document}